# Marginalizing in Undirected Graph and Hypergraph Models

**Enrique F. Castillo**
Dept. of Applied Mathematics
and Computational Sciences.
University of Cantabria
39005 Santander, Spain
e-mail:castie@ccaix3.unican.es

**Juan Ferrándiz**
Department of Statistics
and Operations Research,
University of Valencia,
Dr. Moliner 50, E-46100, SPAIN
e-mail: Juan.Ferrandiz@uv.es

**Pilar Sanmartín**
Department of Mathematics,
Jaume I University
Castellón, SPAIN
e-mail: sanmarti@mat.uji.es

## Abstract

Given an undirected graph $\mathcal{G}$ or hypergraph $\mathcal{H}$ model for a given set of variables $V$, we introduce two marginalization operators for obtaining the undirected graph $\mathcal{G}_A$ or hypergraph $\mathcal{H}_A$ associated with a given subset $A \subset V$ such that the marginal distribution of $A$ factorizes according to $\mathcal{G}_A$ or $\mathcal{H}_A$, respectively. Finally, we illustrate the method by its application to some practical examples. With them we show that hypergraph models allow defining a finer factorization or performing a more precise conditional independence analysis than undirected graph models.

## 1 INTRODUCTION

In many practical situations the structural relationship among a set of variables $V = \{V_1, \ldots, V_n\}$ can be represented as an undirected graph $\mathcal{G} = (V, E)$, where $E$ is the set of edges of $\mathcal{G}$. If two variables are independent, the corresponding nodes should not be connected by a path.

Similarly, if the independence between variables $X$ and $Y$ is indirect and mediated by a third variable $Z$ (that is, if $X$ and $Y$ are conditionally independent given $Z$), we display $Z$ as a node that intersects the path between $X$ and $Y$, i.e., $Z$ is a *cutset* separating $X$ and $Y$. This correspondence between conditional independence and cutset separation in undirected graphs forms the basis of the theory of *Markov fields* (Isham [5], Lauritzen [6], Wermuth and Lauritzen [10]), and has been given axiomatic characterizations (Pearl and Paz [11]).

However, in many practical cases we can be interested not in the whole set of variables $V$ but in a subset $A$ of them. In this case the initial graph model is not the most appropriate to work with and we are interested in the graph model induced by the initial graph in $A$.

The independence graph of marginal probability distributions for a subset of the considered variables was undertaken in Frydenberg (1990), after Asmussen (1983). There, he stated the collapsibility condition for the corresponding subgraph to be the independence graph of the marginal probability distribution.

Unfortunately, not all probabilistic models can be represented by undirected perfect maps. Pearl and Paz [11] characterize the dependency models represented by undirected perfect maps. The theorem refers not only to probabilistic but to general dependency models.

Since the resulting independence graph reveals this lack of sensitivity to detect all independence properties and lack identification of missing $n$-th ($n > 2$) order interactions when second order interactions are present, as an alternative, we use hypergraph models (see Rose [12], Tarjan and Yannakakis [13], Mellouli [9], Studeny [16] and Shafer and Shenoy [15] for related problems). In this paper, based on the factorization properties, we give an algorithm for obtaining the marginal independence graph under general conditions. To illustrate these concepts, we use some examples in which this lack of sensitivity and the characteristic contribution of hypergraph models become apparent.

In Section 2 we introduce the main concepts to be used in the rest of the paper with a distinction between those required for the case of graphs and those for hypergraphs. In particular, we introduce the hypergraph models based on Gibbs distributions. In Section 3 we introduce a marginalization operator for the case of undirected graphs that allows obtaining such a graph in the sense of the marginal model to satisfy the corresponding factorization properties. We also give an algorithm to implement this operator. In Section 4 we follow exactly the same process for the case of hypergraphs. In both sections we illustrate the methods by means of practical examples. Finally, we make some comparisons, and in Section 6 we give some conclusions and recommendations.

## 2 BACKGROUND

We divide this section in two parts. The first is devoted to undirected graphs, and the second to Gibbs distri-butions and hypergraphs. We assume that the range of every variable is a real set containing the zero.

## 2.1 UNDIRECTED GRAPHS

The main theorem to be given in Section 3 requires several concepts of undirected graphs which are given below. We illustrate them with some examples.

**Definition 1 (Path).** *Given a graph $\mathcal{G}$ a path of length $n$ between nodes $V_r$ and $V_s$ is a sequence of nodes $V_0, \ldots, V_n$ such that $(V_i, V_{i+1}); i = 0, \ldots, n-1$ are edges of $\mathcal{G}$ and $V_0 = V_r$ and $V_n = V_s$.*

**Definition 2 (Connected Nodes).** *Given a graph $\mathcal{G} = (V, E)$, two nodes $V_r, V_s \in V$ are said to be connected if there is a path from $V_r$ to $V_s$. They are said to be directly connected iff the path is of length 1.*

**Definition 3 (Complete Set).** *Given a graph $\mathcal{G} = (V, E)$, a set $A \subseteq V$ is said to be complete if all nodes in $A$ are mutually and directly connected by edges in $E$.*

**Definition 4 (Clique).** *A maximal complete set of nodes is called a clique.*

**Definition 5 (Boundary).** *Given a graph $\mathcal{G} = (V, E)$ and a subset $A \subset V$ the boundary $bd(A)$ of $A$ is the set of nodes $V_r \notin A$ such that they are directly connected to an element of $A$, i.e.,*

$$bd(A) = \{V_r \notin A | (V_r, V_s) \in E \text{ or } (V_s, V_r) \in E, V_r \in A\}.$$

**Definition 6 (Connectivity Components).** *Given a graph $\mathcal{G} = (V, E)$ its set of nodes $V$ can be partitioned in maximal subsets of nodes which are mutually connected (see Lauritzen (1996), page 6). These sets are called connectivity components of $\mathcal{G}$.*

**Example 1** *Consider the set of variables $V = \{V_1, V_2, \ldots, V_{10}\}$ and the graph $\mathcal{G} = (V, E)$ shown in Figure 1, where*

$$E = \{(V_1, V_3), (V_1, V_4), (V_1, V_5), (V_2, V_4), (V_3, V_4), (V_3, V_5), (V_5, V_4), (V_6, V_4), (V_7, V_9), (V_7, V_{10}), (V_8, V_{10})\}.$$

*Some illustrative examples of the above definitions are:*

<u>Path</u>: *The sequence of nodes $\{V_1, V_4, V_5, V_3\}$ is a path of length 3 between $V_1$ and $V_3$, as it is the sequence $\{V_1, V_3\}$, which has length 1.*

<u>Connected nodes</u>: *The nodes $V_8$ and $V_9$ are connected nodes because there is a path $\{V_8, V_{10}, V_7, V_9\}$ joining $V_8$ and $V_9$.*

<u>Directly connected nodes</u>: *Nodes $V_7$ and $V_{10}$ are directly connected nodes because the path $\{V_7, V_{10}\}$ joining them has length 1.*

<u>Complete Sets</u>: *The only complete set of four elements in $\mathcal{G}$ is $\{V_1, V_3, V_4, V_5\}$ (all pairs of nodes are directly connected). Obviously, all its subsets are also complete and it contains the only four complete sets of three elements. The remaining complete sets contain one or two elements.*

<u>Clique</u>: *The sets $\{V_1, V_3, V_4, V_5\}$, $\{V_4, V_2\}$, $\{V_4, V_6\}$, $\{V_7, V_9\}$, $\{V_7, V_{10}\}$, $\{V_8, V_{10}\}$ are the cliques of $\mathcal{G}$.*

<u>Boundary Set</u>: *The boundary of the set $\{V_1, V_3, V_4, V_5\}$ is the set $\{V_2, V_6\}$.*

<u>Connectivity components</u>: *The connectivity components of the graph $\mathcal{G}$ are*

$$\tau_1 = \{V_1, V_2, V_3, V_4, V_5, V_6\} \text{ and } \tau_2 = \{V_7, V_8, V_9, V_{10}\}.$$

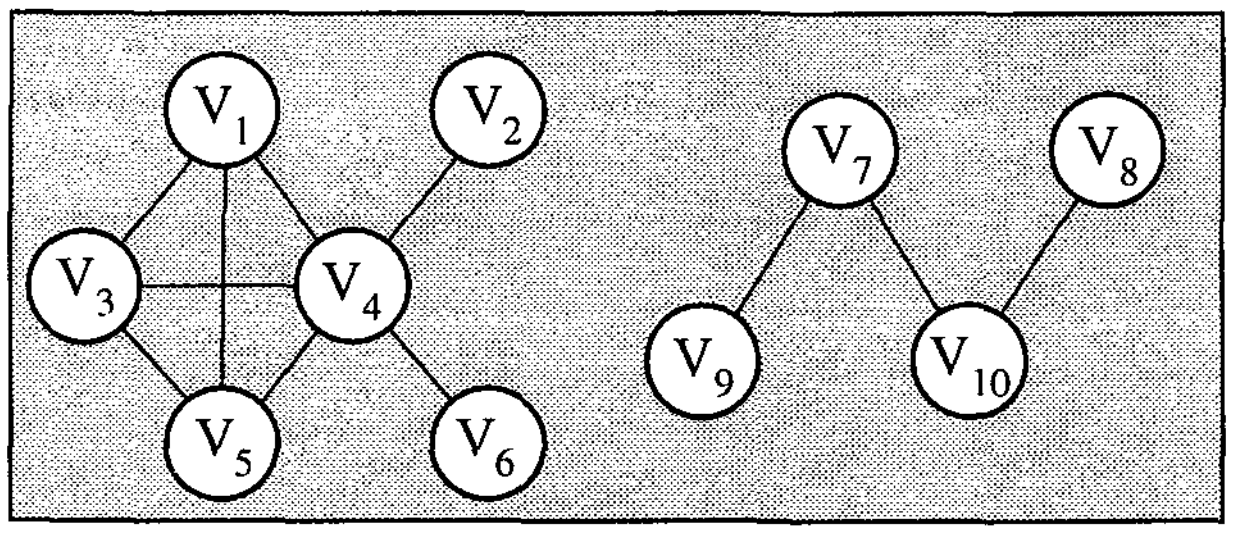

Figure 1: Undirected graph.

**Definition 7 (Completed Edge Set).** *Given a graph $\mathcal{G} = (V, E)$ and a subset $A \subset V$, the completed edge sets $E^*(A)$ of $A$ is the set of all possible edges between nodes in $A$.*

**Definition 8 (Subgraph).** *Given a graph $\mathcal{G} = (V, E)$ and a subset $A \subset V$, the subgraph $\mathcal{G}_A$ is the graph $\mathcal{G}_A = (A, E|_A)$, that is, the graph defined over $A$ and containing the edges of $E$ connecting nodes in $A$.*

**Definition 9 (Factorization Property).** *A probability distribution $P$ on $V$, is said to factorize according to an undirected graph $\mathcal{G}$ (UDG), if for all complete set, $C$, of vertices there exist non-negative functions $\psi_C$ such that*

$$p(v) = \prod_{C \subset V \text{ complete}} \psi_C(c)$$

The above factorization can be done using only cliques. However, this leads to a coarser factorization.

**Example 2** *Consider again the graph in Example 1.*

<u>Completed edge set</u>: *The completed edge set of the set $\{V_7, V_8, V_9\}$ is $\{(V_7, V_8), (V_7, V_9), (V_8, V_9)\}$.*

<u>Subgraph</u>: *The subgraph associated with the set $\{V_2, V_4, V_5, V_6\}$ is*

$$\{\{V_2, V_4, V_5, V_6\}, \{(V_2, V_4), (V_4, V_5), (V_4, V_6)\}.$$

<u>Factorization</u>: *A possible factorization of $p(v)$ is*

$$p(v) = \psi(v_1, v_3, v_4, v_5)\psi(v_2, v_4)\psi(v_4, v_6)\psi(v_7, v_9) \psi(v_7, v_{10})\psi(v_8, v_{10}).$$

## 2.2 GIBBS DISTRIBUTIONS AND HYPERGRAPHS

As it is well known undirected graphs do not lead to the finest possible factorization in probabilistic models. This justifies the use of the Gibbs and hypergraph models to be given below.

**Definition 10 (Gibbs Model).** *Given a graph $\mathcal{G} = (V, E)$, the set of random variables $V$ is said to follow a Gibbs model according to the graph $\mathcal{G}$ if its associated probability density function (pdf) can be written in the form*

$$p(v) = \exp\left(-\sum_{C\in\mathcal{C}} U_C(c)\right)/K, \qquad (1)$$

*where $K$ is a normalizing constant and $\mathcal{C}$ is the class of all complete sets of $V$ with respect to $\mathcal{G}$. The functions $U_C$ are called interaction functions and some of them can be null. (In order to avoid trivial undeterminations we will assume hereafter $U_\emptyset(\cdot) = 0$).*

*The set $U = \{U_C(c)|C \in \mathcal{C}\}$ in (1) is called a potential.*

Note that Expression (1) shows a characteristic factorization property of the corresponding Gibbs model. In fact the density in (1) factorizes as

$$p(v) = \frac{1}{K}\prod_{C\in\mathcal{C}} \exp(-U_C(c)) = \frac{1}{K}\prod_{C\in\mathcal{C}} \psi_C(c), \qquad (2)$$

where the factors in $\{\psi_C(c)|C \in \mathcal{C}\}$ are positive.

The above interpretation of the joint density in terms of the interaction functions is not unique. However, we are interested in the simplest possible representation, which is given by the normalized potential. In it an interaction function $U_C(c)$ appears iff it cannot be written in terms of a sum of functions with less arguments.

**Definition 11 (Normalized Potential).** *A potential $U$ such that $U_C(c) = 0$ whenever some component of $c$ is null is called a normalized potential.*

It can be shown that this potential is unique (see Winkler (1995)). In addition, any given potential $U^0$ can be normalized in the sense of leading to the same joint distribution for $V$, by means of

$$U_C(c) = \sum_{B\subseteq C\subseteq D\subseteq V} (-1)^{|C\setminus B|} U^0_D(b, 0_{D\setminus B}) \qquad (3)$$

This last equation makes evident that the normalized potential produces a finer factorization (2) of the pdf, because for every non-null interaction function $U_C(c)$ of the normalized potential there is at least one non-null interaction function $U^0_D(d)$ involving a bigger set of variables.

**Definition 12 (Potential Restricted to a Set).** *Given a potential $U$ on the set $V$ and a subset $A \subset V$ the potential $U|_A$ restricted to $A$ is the set*

$$U|_A = \{U_C \mid U_C \in U \text{ and } C \subset A\}.$$

**Example 3** *Consider the set of variables $V = \{V_1, V_2, V_3, V_4, V_5, V_6\}$ and the graph $\mathcal{G} = (V, E)$, where*

$$E = \{(V_1, V_2), (V_1, V_3), (V_2, V_3), (V_4, V_5), (V_5, V_6)\}$$

*Gibbs Model: Let us assume the following density:*

$$p(v) \propto \exp(-\theta_{12}(1 + v_1)v_2 - \theta_{13}v_1v_3 - \theta_{23}v_2v_3 - \theta_{45}v_4v_5 - \theta_{56}v_5v_6), \qquad (4)$$

*with associated potential $U^0$:*

$$\{\theta_{12}(1 + v_1)v_2, \theta_{13}v_1v_3, \theta_{23}v_2v_3, \theta_{45}v_4v_5, \theta_{56}v_5v_6\}.$$

*where $\theta_{ij}$ are constants.*

*Normalized Potential: The corresponding normalized potential $U$ becomes:*

$$\{\theta_{12}v_2, \theta_{12}v_1v_2, \theta_{13}v_1v_3, \theta_{23}v_2v_3, \theta_{45}v_4v_5, \theta_{56}v_5v_6\}. \qquad (5)$$

*Potential Restricted to a Set: Given the set $A = \{V_1, V_3, V_5\}$, the potential restricted to $A$ is:*

$$U|_A = \{\theta_{13}v_1v_3\}.$$

**Definition 13 (Hypergraph).** *Given a set $V$, an hypergraph is a subset of parts of $V$.*

**Definition 14 (Hypergraph associated with a family of potentials. Hypergraph Models).** *Given a parametric family of potentials, the hypergraph associated with its normalized potential $U^\theta$ is defined as the class of all sets of $V$ with non-null interaction function $U^\theta_C$ for at least one element in the family, i.e.:*

$$\mathcal{H} = \{C \subseteq V \mid U^\theta_C \not\equiv 0 \text{ for some } \theta\}. \qquad (6)$$

*The corresponding model is called an interaction functions hypergraph or simply hypergraph model.*

Note that hypergraph models are more capable to distinguish models than undirected graph models. For example, the last models cannot distinguish between the hypergraph model with potential (5) and the hypergraph model with potential

$$U_1 = \{\theta_{12}v_2, \theta_{12}v_1v_2, \theta_{13}v_1v_3, \theta_{123}v_1v_2v_3, \theta_{23}v_2v_3, \theta_{45}v_4v_5, \theta_{56}v_5v_6\}. \qquad (7)$$

Every hypergraph $\mathcal{H}$ on $V$ induces in $V$ the graph $\mathcal{G}(\mathcal{H}) = (V, E)$, where

$$E = \{(V_r, V_s) \mid \{V_r, V_s\} \subseteq A \in \mathcal{H}\}.$$

The graph $\mathcal{G}(\mathcal{H})$ associated with the hypergraph of a family of potentials verifies the factorization property with every probability distribution induced by these potentials.

**Definition 15 (Hypergraph Partial Ordering).** *Given two hypergraphs $\mathcal{H}_1$ and $\mathcal{H}_2$ on $V$, we say that $\mathcal{H}_1$ precedes $\mathcal{H}_2$ iff every element of $\mathcal{H}_1$ is contained in an element of $\mathcal{H}_2$, that is,*

$$\mathcal{H}_1 \preceq \mathcal{H}_2 \Leftrightarrow \forall H_1 \in \mathcal{H}_1\ \exists H_2 \in \mathcal{H}_2 \text{ with } H_1 \subseteq H_2$$

Comparing again the potentials in (5) and (7), we can say that the hypergraph associated with (5) precedes the hypergraph associated with (7), but not conversely.

Now we can state the property of normalized potentials producing finer factorizations in the more precise terms of partial ordering of the associated hypergraphs. The hypergraph associated with the normalized potential precedes the hypergraph associated with any other potential leading to the same probability distribution.

**Definition 16 (Boundary Hypergraph).** *Let $\mathcal{H}$ be the hypergraph and $A \subset V$. The boundary hypergraph $\mathcal{H}^A$ of $V \setminus A$ is the hypergraph of all subsets of $A$ which are the boundary of some connectivity component of $\mathcal{G}_{V\setminus A}$ in $\mathcal{G}(\mathcal{H})$.*

**Example 4** *Consider again Example 3.*

*Hypergraph associated with a family of potentials: The hypergraph associated with the potential U is*

$$\mathcal{H} = \{\{V_2\},\{V_1,V_2\},\{V_1,V_3\}, \{V_2,V_3\},\{V_4,V_5\},\{V_5,V_6\}\}.$$

*Graph associated with a hypergraph: The graph associated with hypergraph $\mathcal{H}$ is*

$$(\{V_1,V_2,V_3,V_4,V_5,V_6\}, \{(V_1,V_2),(V_1,V_3),(V_2,V_3),(V_4,V_5),(V_5,V_6)\}).$$

*Boundary Hypergraph: Given $A = \{V_1, V_3, V_5\}$, since the connectivity components of $V \setminus A$ are*

$$\tau_1 = \{V_2\}, \tau_2 = \{V_4\}, \tau_3 = \{V_6\},$$

*the boundary hypergraph $\mathcal{H}^A$ of $V\setminus A$ is the hypergraph:*

$$\{\{V_1,V_3\},\{V_5\}\}.$$

## 3 THE MARGINAL OPERATOR FOR UNDIRECTED GRAPHS

**Theorem 1 (The marginalization theorem for undirected graphs).** *Let $\mathcal{G}$ be the undirected graph $(V,E)$, and $P$ the probability distribution over $V$. If $A \subset V$ and $P_A$ is the marginal distribution associated with $A$, we have that if $P$ factorizes according to the graph $\mathcal{G}$, then, the marginal distribution $P_A$ factorizes according to the graph $\mathcal{G}_A^{ma} = (A, E_A^{ma})$, where*

$$E_A^{ma} = E|_A \cup_{\tau\in\mathcal{T}} E^*(bd(\tau)),$$

*and $\mathcal{T}$ is the set of connectivity components of $\mathcal{G}_{V\setminus A}$.*

**Proof:** The marginal distribution is obtained by integration over de range of $Z = V \setminus A$, that is:

$$p_A(a) = \int p(a,z)dz. \quad (8)$$

Replacing the value of $p$ in terms of its factors and assuming that $C$ varies in the class of all complete sets $\mathcal{C}$, we get:

$$\int \prod_C \psi_C(c)dz = \int \prod_{C\subseteq A} \psi_C(c) \prod_{C\not\subseteq A} \psi_C(c)dz = \prod_{C\subseteq A} \psi_C(c) \int \prod_{C\not\subseteq A} \psi_C(c)dz.$$

Thus,

$$p_A(a) = \psi_{C_0}(c_0) \prod_{C\subseteq A} \psi_C(c),$$

where

$$\psi_{C_0}(c_0) = \int \prod_{C\not\subseteq A} \psi_C(c)dz;\ C_0 = (\cup_{C\not\subseteq A} C)\cap A. \quad (9)$$

Let $\mathcal{T}$ be the set of connectivity components of the subgraph $\mathcal{G}_{V\setminus A}$. Obviously, there are no elements in $\mathcal{C}$ with indices in more than one of these different components. Thus, the integration over $V\setminus A$ in (9) factorizes in integrals, each on a connectivity component, as:

$$\psi_{C_0}(c_0) = \prod_{\tau\in\mathcal{T}} \int \prod_{C\cap\tau\neq\emptyset} \psi_C(c)d\tau,$$

where each factor is of the form:

$$\psi^{\tau}_{bd(\tau)}(bd(\tau)) = \int \prod_{C\cap\tau\neq\emptyset} \psi_C(c)d\tau\,, \quad (10)$$

a function of the set of locations in $A$ which are neighbors of some location in the connectivity component $\tau$, that is, the set $bd(\tau)$. Then, $C_0 = \bigcup_{\tau\in\mathcal{T}} bd(\tau)$.

We shall write (9) as:

$$\psi_{C_0}(c_0) = \prod_{\tau} \psi^{\tau}_{bd(\tau)}(bd(\tau)). \quad (11)$$

Consequently, the marginal pdf can be written as:

$$p_A(a) = \prod_{C\subseteq A} \psi_C(c) \prod_{\tau} \psi^{\tau}_{bd(\tau)}(bd(\tau))\,, \quad (12)$$

where we can see that the distribution $P_A$ satisfies the factorization property with respect to $\mathcal{G}_A^{ma} = (A, E_A^{ma})$, as was to be proven. ∎

This operator reminds us, in a certain way, the moralization of chain graphs, the difference being that this applies to chain graphs (with the existence of arrows) to obtain a directed graph, by "marrying" the parents of each chain component. This new operator applies to undirected graphs and what get married are the elements in the boundaries of the connectivity components of the locations associated with variables disappearing during the marginalization process.

The above theorem suggests the following algorithm for marginalization.

**Algorithm 1 Marginalization**

**Input:** A graph $\mathcal{G} = (V, E)$ and a subset $A \subset V$.

**Output:** A graph $\mathcal{G}^{ma}_A = (A, E^{ma}_A)$ such that the $A$-marginal of the graphical model associated with the graph $\mathcal{G}$ factorizes according to $\mathcal{G}^{ma}_A$.

**Step 1:** Obtain the set $E|_A$ (edges in $\mathcal{G}_A$).

**Step 2:** Obtain the subgraph $\mathcal{G}_{V\setminus A}$.

**Step 3:** Obtain connectivity components $\mathcal{T}$ of $\mathcal{G}_{V\setminus A}$.

**Step 4:** Determine the set $bd(\tau)$ in $\mathcal{G}$ for each $\tau \in \mathcal{T}$.

**Step 5:** Obtain the completed edge sets $E^*(bd(\tau))$ for each $\tau \in \mathcal{T}$.

**Step 6:** Return the graph $\mathcal{G}^{ma}_A = (A, E^{ma}_A)$ where $E^{ma}_A$ is the union of $E|_A$ and $\cup_{\tau\in\mathcal{T}} E^*(bd(\tau))$. ∎

**Example 5** *Assume the graph* $\mathcal{G} = (V, E)$, *where*

$$V = \{V_1, V_2, V_3, V_4, V_5, V_6\}$$
$$E = \{(V_1, V_2), (V_2, V_3), (V_4, V_5), (V_5, V_6)\}$$

*and the set* $A = \{V_1, V_3, V_5\}$.

*If we apply the Algorithm 1, we obtain:*

**Step 1:** $E|_A = \emptyset$.

**Step 2:** $\mathcal{G}_{V\setminus A} = \{\{V_2, V_4, V_6\}, \emptyset\}$.

**Step 3:** $\mathcal{T} = \{\tau_1, \tau_2, \tau_3\} = \{\{V_2\}, \{V_4\}, \{V_6\}\}$.

**Step 4:** $bd(\tau_1) = \{V_1, V_3\}$, $bd(\tau_2) = \{V_5\}$, $bd(\tau_3) = \{V_5\}$.

**Step 5:** $E^*(bd(\tau_1)) = \{(V_1, V_3)\}$, $E^*(bd(\tau_2)) = \emptyset$, $E^*(bd(\tau_3)) = \emptyset$.

**Step 6:** *We return the graph* $\mathcal{G}^m_A = (A, E^m_A) = \{\{V_1, V_3, V_5\}, \{(V_1, V_3)\}\}$.

*Assume that now we add the edge* $(V_1, V_3)$ *to* $E$. *If we apply the Algorithm 1, we obtain:*

**Step 1:** $E|_A = \{\{V_1, V_3\}\}$.

**Step 2:** $\mathcal{G}_{V\setminus A} = \{\{V_2, V_4, V_6\}, \emptyset\}$.

**Step 3:** $\mathcal{T} = \{\tau_1, \tau_2, \tau_3\} = \{\{V_2\}, \{V_4\}, \{V_6\}\}$.

**Step 4:** $bd(\tau_1) = \{V_1, V_3\}$, $bd(\tau_2) = \{V_5\}$, $bd(\tau_3) = \{V_5\}$.

**Step 5:** $E^*(bd(\tau_1)) = \{(V_1, V_3)\}$, $E^*(bd(\tau_2)) = \emptyset$, $E^*(bd(\tau_3)) = \emptyset$.

**Step 6:** *We return the graph* $\mathcal{G}^m_A = (A, E^m_A) = \{\{V_1, V_3, V_5\}, \{(V_1, V_3)\}\}$. ∎

*Note that in both cases we obtain the same marginal graph.*

## 4 THE MARGINAL OPERATOR FOR HYPERGRAPHS

In this section we analyze the marginalization problem in hypergraphs models.

To this aim we use the following theorem, where we state the potential $U^A$ corresponding to the family of marginal distributions $P^\theta_A$ in terms of the changes suffered by the original potential $U$ restricted to the set $A$. From the proof of theorem 1 we have seen the role of the connectivity components $\tau$ of the subgraph $\mathcal{G}_{V\setminus A}$. In fact, we could call their contributions to the marginal potential $U^A$ the *innovations* of the potential $U$. They can be computed for each $B \subseteq A$ as the double sum

$$V^A_B(b) = \sum_{B\subseteq D\in\mathcal{H}^A} \sum_{E\subseteq B} (-1)^{|B\setminus E|} U^*_D(e, 0_{D\setminus E}) \quad (13)$$

if there is $D$ in $\mathcal{H}^A$ including $B$, $(B \neq \emptyset)$, and 0 otherwise, where

$$U^*_D(d) = \sum_{\tau : bd(\tau)=D} U^\tau_{bd(\tau)}(d), \quad (14)$$

with

$$U^\tau_{bd(\tau)}(d) = -\ln\Big(\int \exp\Big(-\sum_{C: C\cap\tau\neq\emptyset} U_C(c)\Big)\Big)dv_\tau. \quad (15)$$

Now, we shall state without proof the following theorem (Sanmartín (1997)) for the parametric potential $U^\theta$ of a hypergraph model, where we add the $\theta$ superscript to all the functions derived from $U^\theta$ whenever we want to emphasize their parametric character.

**Theorem 2 (The marginalization theorem for interaction hypergraphs).** *Consider a parametric family of Gibbs models over* $V$, *with interaction functions hypergraph* $\mathcal{H}$, *and* $P^\theta_A$ *be the corresponding family of marginal distributions over* $A$. *Then, the interaction function hypergraph* $\mathcal{H}_A$ *of the family* $P^\theta_A$ *can be expressed as:*

$$\mathcal{H}_A = (\mathcal{H}|_A \cup \mathcal{H}^+_A) \setminus \mathcal{H}^-_A, \quad (16)$$

*where*

1. $\mathcal{H}|_A$ *is the restriction of* $\mathcal{H}$ *to* $A$, *that is, the set of elements in* $\mathcal{H}$ *which are subsets of* $A$.

2. $\mathcal{H}^+_A$ *is the family of subsets* $B \subseteq A$ *not in* $\mathcal{H}|_A$ *and such that* ${}^\theta V^A_B(b)$ *is a non-null function for some* $\theta$. *(These are the new complete sets that will appear after marginalization).*

3. $\mathcal{H}^-_A$ *is the set of complete sets in* $\mathcal{H}|_A$ *such that they are subsets of some set in* $\mathcal{H}^A$ *and satisfy the equation:*

$$U^\theta_B(b) = -{}^\theta V^A_B(b); \ \forall\theta \quad (17)$$

*(These are the complete sets that will disappear after marginalization).*

The decomposition of $\mathcal{H}_A$ above, exhibits the necessary and sufficient conditions for graphical and parametric collapsibility. It becomes apparent from (16) that the graph $\mathcal{G}(\mathcal{H}_A)$ associated with the marginal hypergraph coincides with the subgraph $\mathcal{G}(\mathcal{H})_A \equiv \mathcal{G}(\mathcal{H}|_A)$ iff $\mathcal{H}_A^+ \preceq \mathcal{H}|_A$ and ($\mathcal{H}_A^- = \emptyset$ or $\mathcal{H}_A^- \preceq (\mathcal{H}|_A \setminus \mathcal{H}_A^-)$).

If, in addition, we require parametric collapsibility ($U|_A = U^A$), that is, the marginalizing operation not to change the interaction functions involving variables in $A$, the necessary and sufficient condition becomes that all innovations $V_B^A(b)$ in (13) be null.

This theorem suggests the algorithm below for marginalizing a hypergraph. In it we clarify the meaning of the sets and functions appearing in Expressions (13) and (15), which are not easy to understand. With the same purpose we also include a simple example.

**Algorithm 2 Marginalization of Hypergraphs.**

- **Input:** A set $V$, a parametric family of normalized potentials $U^\theta$ over $V$, and a subset $A \subset V$.
- **Output:** The $A$-marginal potential ${}^\theta U^A$, together with its associated hypergraph $\mathcal{H}_A$ and graph $\mathcal{G}(\mathcal{H}_A)$.

**Step 1:** Obtain the hypergraph $\mathcal{H}$ associated with the given potential $U$.

**Step 2:** Obtain the graph $\mathcal{G}(\mathcal{H})$ associated with the hypergraph.

**Step 3:** Determine the connectivity components of the subgraph associated with $V \setminus A$.

**Step 4:** Obtain the boundary hypergraph $\mathcal{H}^A$, as the collection of the boundaries in $\mathcal{G}(\mathcal{H})$ of the connectivity components of $V \setminus A$.

**Step 5:** For each element $B \in \mathcal{H}^A$ and each $\tau$ verifying $bd(\tau) = B$ in (15) calculate the functions $U_B^\tau(b)$.

**Step 6:** For each element $B \in \mathcal{H}^A$ calculate the functions $U_B^*(b)$ (see (14)).

**Step 7:** Using (13), calculate $V_B^A(b)$ for each non-void subset $B$ of the sets in $\mathcal{H}^A$.

**Step 8:** Calculate the A-marginal potencial $U^A$ by adding $V^A$ to the initial potential $U$ restricted to $A$.

**Step 9:** Obtain the hypergraph $\mathcal{H}_A$ associated with $U^A$.

**Step 10:** Obtain the graph $\mathcal{G}(\mathcal{H}_A)$ associated with $U^A$.

**Step 11:** Return $U^A$, $\mathcal{H}_A$ and $\mathcal{G}(\mathcal{H}_A)$. ∎

**Example 6** *Assume the set*

$$V = \{V_1, V_2, V_3, V_4, V_5, V_6\},$$

*of binary* (0, 1) *variables, the normalized potential*

$$U = \{\alpha_{12}v_2, \alpha_{12}v_1v_2, \alpha_{23}v_2v_3, \alpha_{45}v_4v_5, \alpha_{56}v_5v_6\}$$

*with*

$$\alpha_{12} \neq 0; \alpha_{23} \neq 0; \alpha_{45} \neq 0; \alpha_{56} \neq 0 \tag{18}$$

*and the* $V$*-subset* $A = \{V_1, V_3, V_5\}$.

**Step 1:** *The hypergraph* $\mathcal{H}$ *associated with the given potential* $U$ *is*

$$\mathcal{H} = \{\{V_2\}, \{V_1, V_2\}, \{V_2, V_3\}, \{V_4, V_5\}, \{V_5, V_6\}\}.$$

**Step 2:** *The graph associated with the hypergraph is given in Figure 2(a)*

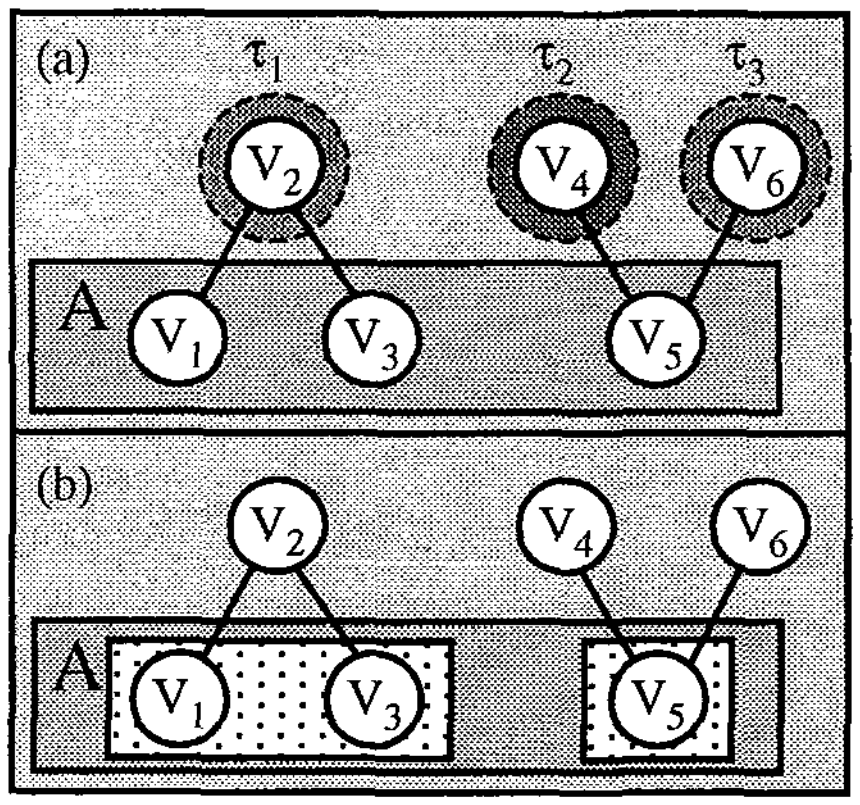

Figure 2: (a) Graph associated with the hypergraph in Example 6 showing the connectivity components $\tau_1, \tau_2, \tau_3$ of the subgraph associated with $V \setminus A$, and (b) boundary hypergraph $\mathcal{H}^A$ (sets in dotted regions).

**Step 3:** *The connectivity components of the subgraph associated with* $V \setminus A$ *are*

$$\tau_1 = \{V_2\}; \tau_2 = \{V_4\} \text{ and } \tau_3 = \{V_6\};$$

*as it can be seen in Figure 2(a).*

**Step 4:** *Since the boundaries of the connectivity components are* $bd(\tau_1) = \{V_1, V_3\}$, $bd(\tau_2) = \{V_5\}$ *and* $bd(\tau_3) = \{V_5\}$, *the boundary hypergraph is*

$$\mathcal{H}^A = \{\{V_1, V_3\}, \{V_5\}\},$$

*which is shown in Figure 2(b) (sets in dotted regions).*

**Step 5:** *For the first element* $\{V_1, V_3\}$ *(see (15)):* $U_{\{V_1,V_3\}}^{\tau_1} =$

$$-\ln(\sum_{v_2=0}^{v_2=1} \exp\{-\alpha_{12}v_1v_2 - \alpha_{12}v_2 - \alpha_{23}v_2v_3\})$$

*and for the second* $\{V_5\}$*:*

$$U_{\{V_5\}}^{\tau_2} = -\ln(\sum_{v_4=0}^{v_4=1} \exp\{-\alpha_{45}v_4v_5\}),$$

$$U_{\{V_5\}}^{\tau_3} = -\ln(\sum_{v_6=0}^{v_6=1} \exp\{-\alpha_{56}v_5v_6\})$$

**Step 6:** *For the first element* $\{V_1, V_3\}$ *we have (see (14)):*

$$U_{\{V_1,V_3\}}^* = U_{\{V_1,V_3\}}^{\tau_1}$$

$$= -\ln(\sum_{v_2=0}^{v_2=1} \exp\{-\alpha_{12}v_1v_2 - \alpha_{12}v_2 - \alpha_{23}v_2v_3\})$$

*and for the second:*

$$\begin{aligned} U^*_{\{V_5\}} &= U^{\tau_2}_{\{V_5\}} + U^{\tau_3}_{\{V_5\}} \\ &= -\ln(\sum_{v_4=0}^{v_4=1} \exp\{-\alpha_{45}v_4v_5\}) \\ &\quad -\ln(\sum_{v_6=0}^{v_6=1} \exp\{-\alpha_{56}v_5v_6\}). \end{aligned}$$

**Step 7:** *Since the non-void subset of the sets in* $\mathcal{H}^A = \{\{V_1, V_3\}, \{V_5\}\}$ *are*

$$\{V_1\}, \{V_3\}, \{V_1, V_3\}, \{V_5\},$$

*we have:*

- *For* $\{V_1\}$: $V^A_{\{V_1\}}(v_1)$ *is*

$$-\ln(\exp\{-\alpha_{12}v_1 - \alpha_{12}\} + 1) + \ln(\exp(-\alpha_{12}) + 1).$$

- *For* $\{V_3\}$: $V^A_{\{V_3\}}(v_3)$ *is*

$$-\ln(\exp\{-\alpha_{23}v_3 - \alpha_{12}\} + 1) + \ln(\exp\{-\alpha_{12}\} + 1).$$

- *For* $\{V_1, V_3\}$: $V^A_{\{V_1,V_3\}}(v_1, v_3)$ *is*

$$\begin{aligned} &-\ln(\exp\{-\alpha_{12}v_1 - \alpha_{23}v_3 - \alpha_{12}\} + 1) \\ &+\ln(\exp\{-\alpha_{12}v_1 - \alpha_{12}\} + 1) \\ &+\ln(\exp\{-\alpha_{23}v_3 - \alpha_{12}\} + 1) \\ &-\ln\{\exp(-\alpha_{12}) + 1\}. \end{aligned}$$

- *For* $\{V_5\}$: $V^A_{\{V_5\}}(v_5)$ *is*

$$\begin{aligned} &-\ln(\exp\{-\alpha_{45}v_5\} + 1) + \ln 2 \\ &-\ln(\exp\{-\alpha_{56}v_5\} + 1) + \ln 2. \end{aligned}$$

**Step 8:** *Since the potential U restricted to A is void, the potential* $U^A$ *reduces to* $V^A_B(b)$.

*Due to the fact that we are interested in the non-null interaction functions we must check whether the candidate functions are non-null. However, the following equation:*

$$V^A_{\{V_1,V_3\}}(v_1, v_3) \equiv 0$$

*has only the trivial solutions* $\alpha_{12} = 0$ *or* $\alpha_{23} = 0$ *that contradict the assumption (18). Thus,* $U^A$ *becomes:*

$$\{V^A_{\{V_1\}}(v_1), V^A_{\{V_3\}}(v_3), V^A_{\{V_1,V_3\}}(v_1, v_3), V^A_{\{V_5\}}(v_5)\}.$$

**Step 9:** *The marginal hypergraph becomes:*

$$\mathcal{H}_A = \{\{V_1\}, \{V_3\}, \{V_1, V_3\}, \{V_5\}\}.$$

**Step 10:** *Finally, the associated marginal graph is*

$$\mathcal{G}(\mathcal{H}_A) = (\{V_1, V_3, V_5\}, \{(V_1, V_3)\}).$$

**Step 11:** *Return* $U^A$, $\mathcal{H}_A$ *and* $\mathcal{G}(\mathcal{H}_A)$.

*Note that we have obtained the same solution as with the undirected graph algorithm (see Step 6 in Example 5). However, if the potential U includes the interaction function* $\alpha_{13}v_1v_3$, *as follows:*

$$U = \{\alpha_{12}v_2, \alpha_{12}v_1v_2, \alpha_{13}v_1v_3, \alpha_{23}v_2v_3, \alpha_{45}v_4v_5, \alpha_{56}v_5v_6\},$$

*the following steps above suffer changes:*

**Modified Step 1:** *The hypergraph* $\mathcal{H}$ *associated with the given potential U is*

$$\{\{V_2\}, \{V_1, V_2\}, \{V_1, V_3\}, \{V_2, V_3\}, \{V_4, V_5\}, \{V_5, V_6\}\}.$$

**Modified Step 2:** *The graph associated with the hypergraph is given in Figure 3(a). Note that a new edge appears connecting* $V_1$ *and* $V_3$.

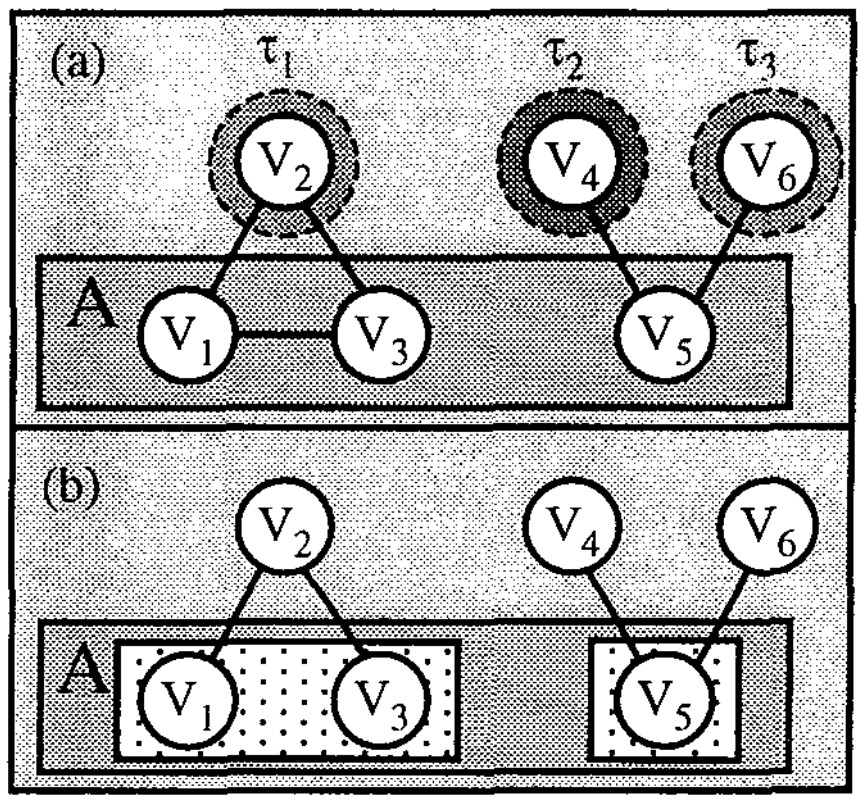

Figure 3: (a) Graph associated with the hypergraph in Example 6 showing the connectivity components $\tau_1, \tau_2, \tau_3$ of the subgraph associated with $V \setminus A$, and (b) boundary hypergraph $\mathcal{H}^A$ (sets in dotted regions).

**Modified Step 8:** *Since we are interested in the non-null interaction functions we must check whether the candidate functions are non-null.*

*We have the following cases:*

**Case 1:** *The interaction function in* $v_1$ *and* $v_3$ *becomes null:*

$$\begin{aligned} &V^A_{\{V_1,V_3\}}(v_1, v_3) + \alpha_{13}v_1v_3 \\ &\quad = -\ln(\exp\{-\alpha_{12}v_1 - \alpha_{23}v_3 - \alpha_{12}\} + 1) \\ &\qquad +\ln(\exp\{-\alpha_{12}v_1 - \alpha_{12}\} + 1) \\ &\qquad +\ln(\exp\{-\alpha_{23}v_3 - \alpha_{12}\} + 1) \\ &\qquad -\ln\{\exp(-\alpha_{12}) + 1\} + \alpha_{13}v_1v_3 \equiv 0 \end{aligned}$$

*which implies*

$$\begin{aligned} \alpha_{13} = &\ln(\exp\{-2\alpha_{12} - \alpha_{23}\} + 1) \\ &-\ln(\exp\{-2\alpha_{12}\} + 1) \\ &-\ln(\exp\{-\alpha_{23} - \alpha_{12}\} + 1) \\ &+\ln\{\exp(-\alpha_{12}) + 1\}. \end{aligned} \quad (19)$$

*Thus, Case* 1 *refers to the family of Gibbs models satisfying (19).*

*Then we have*

$$U^A = \{V^A_{\{V_1\}}(v_1), V^A_{\{V_3\}}(v_3), V^A_{\{V_5\}}(v_5)\}.$$

**Case 2:** *Otherwise, since the potential U restricted to A is* $\{\alpha_{13}v_1v_3\}$, *the potential* $U^A$ *becomes*

$$U^A = \{V^A_{\{V_1\}}(v_1), V^A_{\{V_3\}}(v_3), V^A_{\{V_1,V_3\}}(v_1,v_3) + \alpha_{13}v_1v_3, V^A_{\{V_5\}}(v_5)\}.$$

**Modified Step 9:** *The marginal hypergraph becomes:*

**Case 1:** $\mathcal{H}_A = \{\{V_1\},\{V_3\},\{V_5\}\}$.

**Case 2:** $\mathcal{H}_A = \{\{V_1\},\{V_3\},\{V_1,V_3\},\{V_5\}\}$.

**Modified Step 10:** *Finally, the associated marginal graph is*

**Case 1:** $\mathcal{G}(\mathcal{H}_A) = (\{V_1,V_3,V_5\},\emptyset)$.

**Case 2:** $\mathcal{G}(\mathcal{H}_A) = (\{V_1,V_3,V_5\},\{(V_1,V_3)\})$. ∎

Let $\mathcal{G}^{ma}_A$ and $\mathcal{G}(\mathcal{H}_A)$ be the marginal graph (algorithm 1), and the graph associated with the marginal hypergraph (algorithm 2), for the subset of variables $A$, respectively. Then, from Examples 5 and 6 we can conclude:

1. In case the edge $(V_1,V_3)$ is not in $\mathcal{G} \equiv \mathcal{G}(\mathcal{H})$, then $(V_1,V_3)$ is an edge of both $\mathcal{G}^{ma}_A$ and $\mathcal{G}(\mathcal{H}_A)$.
2. In case the edge $(V_1,V_3)$ is in $\mathcal{G} \equiv \mathcal{G}(\mathcal{H})$, then $(V_1,V_3)$ is an edge of $\mathcal{G}^{ma}_A$, but, if condition (19) applies, it is not and edge of $\mathcal{G}(\mathcal{H}_A)$.

The absence of $(V_1,V_3)$ as an edge of a graph over $A = \{V_1,V_3,V_5\}$ implies the stochastic independence of both variables conditioned to $V_5$. This independence statement is included in the model which has $(V_1,V_3)$ as an edge of $\mathcal{G}(\mathcal{H})$ and verifies (19). This shows that, in this case, the undirected graph representation of the model is not able to capture this separating statement while the hypergraph model is.

## 5 EXAMPLE OF APPLICATION

In this example, the objective is to assess the damage of reinforced concrete structures of buildings. This example, which is taken from Liu and Li (1994) (see also Castillo, Gutiérrez, and Hadi (1997)), is slightly modified for illustrative purposes. The goal variable (the damage of a reinforced concrete beam) is denoted by $X_1$. A civil engineer initially identifies 16 variables $(X_9,\ldots,X_{24})$ as the main variables influencing the damage of reinforced concrete structures. In addition, the engineer identifies seven intermediate unobservable variables $(X_2,\ldots,X_8)$ that define some partial states of the structure. Table 1 shows the list of variables and their definitions.

In our example, the engineer specifies the following cause-effect relationships, as depicted in Figure 4(a). The goal variable $X_1$, is related primarily to three factors: $X_9$, the weakness of the beam available in the form of a damage factor; $X_{10}$, the deflection of the beam; and $X_2$, its cracking state. The cracking state, $X_2$, is related to four variables: $X_3$, the cracking state in the shear domain; $X_6$, the evaluation of the shrinkage cracking; $X_4$, the evaluation of the steel corrosion; and $X_5$, the cracking state in the flexure domain. Shrinkage cracking, $X_6$, is related to shrinkage, $X_{23}$, and the corrosion state, $X_8$. Steel corrosion, $X_4$, is related to $X_8$, $X_{24}$, and $X_5$. The cracking state in the shear domain, $X_3$, is related to four factors: $X_{11}$, the position of the worst shear crack; $X_{12}$, the breadth of the worst shear crack; $X_{21}$, the number of shear cracks; and $X_8$. The cracking state in the flexure domain, $X_5$ is affected by three variables: $X_{13}$, the position of the worst flexure crack; $X_{22}$, the number of flexure cracks; and $X_7$, the worst cracking state in the flexure domain. The variable $X_{13}$ is influenced by $X_4$. The variable $X_7$ is a function of five variables: $X_{14}$, the breadth of the worst flexure crack; $X_{15}$, the length of the worst flexure crack; $X_{16}$, the cover; $X_{17}$, the structure age; and $X_8$, the corrosion state. The variable $X_8$ is related to three variables: $X_{18}$, the humidity; $X_{19}$, the PH value in the air; and $X_{20}$, the content of chlorine in the air.

| $X_i$ | Definition |
|---|---|
| $X_1$ | Damage assessment |
| $X_2$ | Cracking state |
| $X_3$ | Cracking state in shear domain |
| $X_4$ | Steel corrosion |
| $X_5$ | Cracking state in flexure domain |
| $X_6$ | Shrinkage cracking |
| $X_7$ | Worst cracking in flexure domain |
| $X_8$ | Corrosion state |
| $X_9$ | Weakness of the beam |
| $X_{10}$ | Deflection of the beam |
| $X_{11}$ | Position of the worst shear crack |
| $X_{12}$ | Breadth of the worst shear crack |
| $X_{13}$ | Position of the worst flexure crack |
| $X_{14}$ | Breadth of the worst flexure crack |
| $X_{15}$ | Length of the worst flexure cracks |
| $X_{16}$ | Cover |
| $X_{17}$ | Structure age |
| $X_{18}$ | Humidity |
| $X_{19}$ | PH value in the air |
| $X_{20}$ | Content of chlorine in the air |
| $X_{21}$ | Number of shear cracks |
| $X_{22}$ | Number of flexure cracks |
| $X_{23}$ | Shrinkage |
| $X_{24}$ | Corrosion |

Table 1: Definitions of the variables related to damage assessment of reinforced concrete structures.

A graphical representation of the damage problem is shown in Figure 4(a).

Suppose that we are interested in suppressing all the nodes related to the flexion of the beam and keep the remaining nodes (Set $A$), that is (see Figure 4(b)):

$$V \setminus A = \{X_5,X_7,X_{13},X_{14},X_{15},X_{16},X_{17},X_{22},X_{23}\}.$$

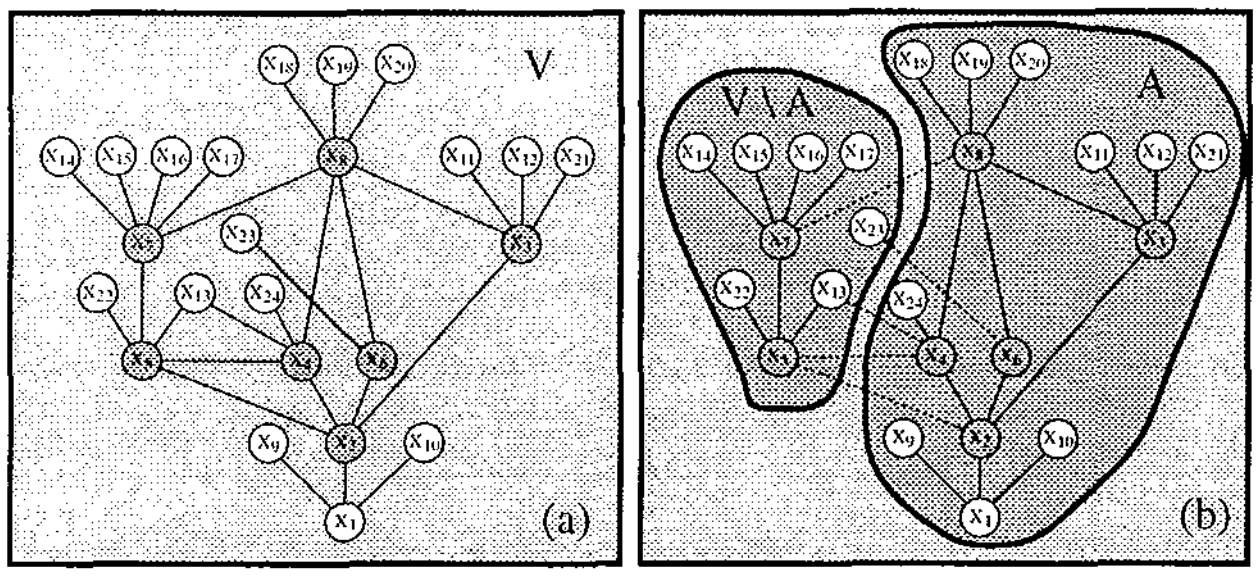

Figure 4: (a) Undirected graph representing the variable relations for the damage assessment of reinforced concrete structure, and (b) sets $A$, $V \setminus A$, and subgraphs associated with $A$ and $V \setminus A$ (continuous edges only).

## 5.1 GRAPH APPROACH

In this case, to marginalize over $A$, we can apply Algorithm 1.

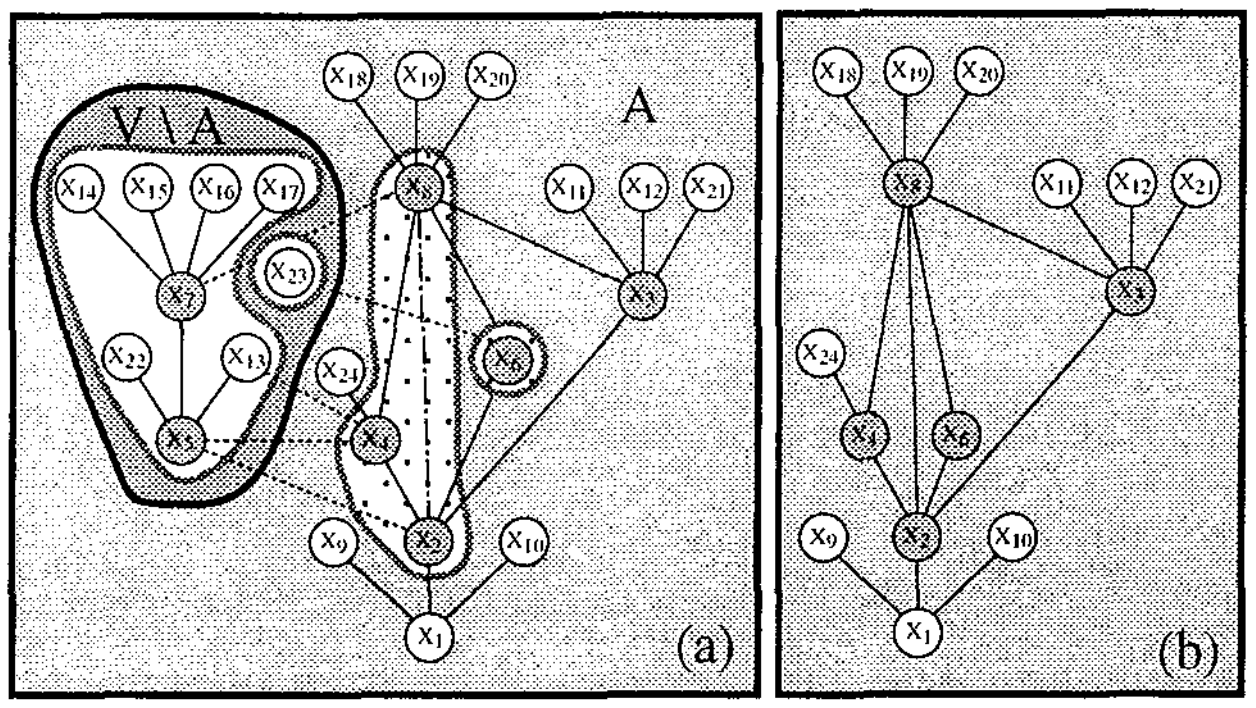

Figure 5: (a) Set $V \setminus A$ with its conectivity components and their completed boundaries (doted regions), and (b) the resulting marginal graph $\mathcal{G}_A^m$ on $A$.

**Step 1:** The set $E|_A$, that is, the set of edges in the subgraph $\mathcal{G}_A$ is shown in Figure 4(b) (the continuous edges in the region $A$).

**Step 2:** The subgraph $\mathcal{G}_{V \setminus A}$ appears in Figure 4(b) (region $A$ with continuous edges).

**Step 3:** The connectivity components $\mathcal{T}$ of $\mathcal{G}_{V \setminus A}$:

$$\begin{aligned} \tau_1 &= \{X_5, X_7, X_{13}, X_{14}, X_{15}, X_{16}, X_{17}, X_{22}, \}; \\ \tau_2 &= \{X_{23}\}, \end{aligned}$$

are shown in Figure 5(a) as white regions.

**Step 4:** The boundaries of the two connectivity components are $bd(\tau_1) = \{X_2, X_4, X_8\}$ and $bd(\tau_2) = \{X_6\}$, as shown in Figure 5(a) where they have been shadowed with dots.

**Step 5:** To complete the set $bd(\tau_1)$ we need to add the edge $(X_2, X_8)$ to the already two existing edges $(X_2, X_4)$ and $(X_4, X_8)$.

**Step 6:** We return the graph in Figure 5(b), which incorporates the edge $(X_2, X_8)$ to the subgraph $\mathcal{G}_A$, thus, showing that the graph $A$ is not collapsible with respect to $A$. ∎

## 5.2 HYPERGRAPH APPROACH

When applying algorithm 2, the differences with the preceding results could only appear in the boundaries of the connectivity components of $V \setminus A$, that is, $bd(\tau_1) = \{X_2, X_4, X_8\}$ and $bd(\tau_2) = \{X_6\}$. The non-null innovations (13) could only arise for subsets of variables contained in these sets. As $bd(\tau_2)$ has only one variable, our problem of exploring possible differences between $\mathcal{G}_A^{ma}$ and $\mathcal{G}(\mathcal{H}_A)$ reduce to those edges connecting variables in $bd(\tau_1)$.

To illustrate, let us assume a Gaussian distribution with mean $\mu$ and dispersion matrix $\Sigma$ for the 24 variables in V. To express this distribution as a hypergraph model, it is easier to work with the precision matrix $\Upsilon = \Sigma^{-1}$. In fact its pdf can be written as

$$\begin{aligned} p(v) &\propto \exp\left(-\tfrac{1}{2}(v-\mu)'\Upsilon(v-\mu)\right) \\ &\propto \exp\left(\sum_i v_i(\tfrac{1}{2}\sum_j \Upsilon_{ij}\mu_j) - \sum_{i \neq j} \tfrac{1}{2}\Upsilon_{ij} v_i v_j\right) \end{aligned} \quad (20)$$

corresponding to expression (1) with normalized potential. Equation (20) shows the relationship between edges in $\mathcal{G}$ and non null elements of the matrix $\Upsilon$.

It is a well known fact that the marginal distribution of a multivariate Gaussian model is again multivariate Gaussian, with precision matrix

$$\Upsilon^A = \Upsilon_{AA} - \Upsilon_{A,V\setminus A}(\Upsilon_{V\setminus A,V\setminus A})^{-1}\Upsilon_{V\setminus A,A}, \quad (21)$$

where the subscripts of $\Upsilon$ stand for the appropriate partition.

Equation (21) shows the decomposition of the precision matrix $\Upsilon^A$ related to the marginal normalized potential $U^A$ in two components:

(i) the matrix $\Upsilon_{A,A}$ corresponding to the restricted potential $U|_A$, and

(ii) the matrix $\Gamma^A = \Upsilon_{A,V\setminus A}(\Upsilon_{V\setminus A,V\setminus A})^{-1}\Upsilon_{V\setminus A,A}$ corresponding to innovations $V_B^A(b)$ of (13).

In particular, the innovation (13) for two variables $V_i$ and $V_j$ in $A$ is

$$\begin{aligned} V^A_{\{V_i,V_j\}}(v_i,v_j) &= -\Gamma^A_{ij} v_i v_j \\ &= -\Big(\sum_{\substack{r,s \in V \setminus A \\ i \sim r,\, j \sim s}} \rho_{rs}\Upsilon_{ir}\Upsilon_{sj}\Big)\, v_i v_j \end{aligned} \quad (22)$$

where $\rho_{rs}$ stands for the $rs$-element of matrix $(\Upsilon_{V\setminus A,V\setminus A})^{-1}$ and $i \sim r$ indicates that node $i$ is directly connected to node $r$ in the associated graph $\mathcal{G}$.

Particularizing to our example, the only edges subject to change when applying Algorithm 2 are $\{(X_2, X_4), (X_2, X_8), (X_4, X_8)\}$.

The edge $(X_2, X_8)$, which was not present in the original graph $\mathcal{G}$, arises as a consequence of the innovation $\Gamma^A_{2,8} = \rho_{5,7}\Upsilon_{2,5}\Upsilon_{7,8}$, and it is null only if $\rho_{5,7}$ vanishes. Matrix $\Upsilon$, being a precision matrix, is definite positive, implying $D = |\Upsilon_{V\setminus A, V\setminus A}| > 0$. After some algebra, $\rho_{5,7}$ can be written as

$$\Upsilon_{13,13}\Upsilon_{14,14}\Upsilon_{15,15}\Upsilon_{16,16}\Upsilon_{17,17}\Upsilon_{22,22}\Upsilon_{2,5}\Upsilon_{5,7}\Upsilon_{7,8}/D$$

which cannot be null unless one or more of the parameteres $\Upsilon_{2,5}$, $\Upsilon_{5,7}$ and $\Upsilon_{7,8}$ vanish. But this would contradict the initial specification of $\mathcal{G}$. Then, the edge $(X_2, X_8)$ will always be present in $\mathcal{G}^{ma}_A$ and $\mathcal{G}(\mathcal{H}_A)$.

Conditions for $(X_2, X_4)$ and $(X_4, X_8)$ to disappear in $\mathcal{G}(\mathcal{H}_A)$ are $\Upsilon_{2,4} = \Gamma_{2,4}$ and $\Upsilon_{4,8} = \Gamma_{4,8}$ , respectively.

They state functional relationships between the parameters $\Upsilon_{2,4}$, $\Upsilon_{4,8}$ and those in $\Upsilon_{V\setminus A, V\setminus A}$. These relationships are compatible with the initial graph $\mathcal{G}$. Thus, the marginal graphs $\mathcal{G}^{ma}_A$ and $\mathcal{G}(\mathcal{H}_A)$ could differ in edges $(X_2, X_4)$ and $(X_4, X_8)$, according to these conditions.

Thus, the example illustrates clearly the advantages of hypergraph models over the usual graph models.

## 6 CONCLUSIONS AND RECOMMENDATIONS

Hypergraph models have been shown to be a powerful alternative to undirected graph models. The main advantage consists of its capability to produce finer factorizations and to catch a more complete set of conditional independence statements. Given a set of variables and an undirected graph or hypergraph model, two algorithms have been given for obtaining the corresponding marginal graph and hypergraph, such that the marginal distribution factorizes according to them. The examples have shown that in some cases the hypergraph is able to capture conditional independence statements that the graph fails to detect. In addition, theorem 2 states a general framework to understand the necessary and sufficient conditions of graphical and parametric collapsibility.

## Acknowledgments

The authors are grateful to Iberdrola, the Universities of Cantabria, Jaume I and Valencia, and the Dirección General de Investigación Científica y Técnica (DGICYT) (projects TIC96-0580 and PB96-0776) for partial support of this research.

## References

[1] Asmussen, S. and Edwards, D., (1983). Collapsibility and response variables in contingency tables. *Biometrika*, 70, 567–578.

[2] E. Castillo, J.M. Gutiérrez, and A.S. Hadi, (1997). *Expert Systems and Probabilistic Network Models.* Springer Verlag, New York.

[3] Darroch, J. N., Lauritzen, S. L., and Speed, T. P. (1980), Markov Fields and Log-linear Models for Contingency Tables. *Annals of Statistics*, 8:522–539.

[4] Frydenberg, M. (1990), Marginalization and collapsibility in graphical interaction models *Annals of Statistics*, 18:790–805.

[5] Isham, B. (1981), An Introduction to Spatial Point Processes and Markov Random Fields. *International Statistical Review*, 49:21–43.

[6] Lauritzen, S. L. (1989), *Lectures on Contingency Tables, 3rd edition.* Aalborg University Press, Aalborg, Denmark.

[7] Lauritzen, S.L. (1996), *Graphical Models.* Oxford University Press, Oxford Statistical Science Series, Vol 17.

[8] Liu, X. and Li, Z. (1994), A Reasoning Method in Damage Assessment of Buildings. *Microcomputers in Civil Engineering, Special Issue on Uncertainty in Expert Systems*, 9:329–334.

[9] Mellouli, K. (1987), On the Propagation of Belief Networks Using the Dempster-Shafer Theory of Evidence. Ph. D. Dissertation, School of Business, University of Kansas.

[10] Wermuth, N. and Lauritzen, S. L. (1983), Graphical and Recursive Models for Contingency Tables. *Biometrika*, 70:537–552.

[11] Pearl, J. and Paz, A. (1987), Graphoids: A Graph-Based Logic for Reasoning about Relevance Relations. In Boulay, B. D., Hogg, D., and Steels, L., editors, *Advances in Artificial Intelligence-II.* North Holland, Amsterdam, 357–363.

[12] Rose, D. J. (1970), Triangulated Graphs and the Elimination Process. Journal of Math. Anal. Appl. 32, 597–609.

[13] Tarjan, R. and Yannakakis, M. (1984), Simple linear-time algorithms to test chordality of graphs, test acyclicity of hypergraphs, and selectively reduce acyclic hypergraphs. SIAM Journal of Computing, 13(3).

[14] Sanmartín, P. (1997), Agregación Temporal en Modelos de Grafos Cadena. Ph. D. Thesis, University of Valencia.

[15] Shafer, G. and Shenoy, P. (1990), Probability Propagation. Annals of Mathematics and Artificial Intelligence, 2: 327–352.

[16] Studeny, M. (1994), On Marginalization, Collapsibility, and precollapsibility. In V. benes, J. Stepan eds.: Distributions with Given Marginals and Moment Problems, Kluwer, Dordrecht, 191–198

[17] Winkler, G. (1995), Image Analysis, Random Fields and Dynamic Monte Carlo Methods. A Matematical Introduction. Springer.